\documentclass[10pt,twocolumn,letterpaper]{article}

\usepackage{cvpr}
\usepackage{times}
\usepackage{epsfig}
\usepackage{graphicx}
\usepackage{amsmath}
\usepackage{amssymb}
\usepackage{bm}
\usepackage{booktabs}
\usepackage{colortbl}
\definecolor{mygray}{gray}{.9}
\usepackage{enumitem}
\usepackage{adjustbox}

\usepackage[breaklinks=true,bookmarks=false]{hyperref}

\cvprfinalcopy 

\def\cvprPaperID{0473} 

\ifcvprfinal\fi
\begin{document}

\title{Unsupervised Image Captioning}

\author{Yang Feng$^\sharp$\thanks{This work was done while Yang Feng was a Research Intern with Tencent AI Lab.} \qquad Lin Ma$^{\natural}$\thanks{Corresponding author.} \qquad Wei Liu$^\natural$ \qquad Jiebo Luo$^\sharp$\\
$^\natural$Tencent AI Lab \quad $^\sharp$University of Rochester\\
{\texttt{\small \{yfeng23,jluo\}@cs.rochester.edu\qquad forest.linma@gmail.com \qquad wl2223@columbia.edu}}
}

\maketitle
\thispagestyle{empty}

\begin{abstract}
Deep neural networks have achieved great successes on the image captioning task. However, most of the existing models depend heavily on paired image-sentence datasets, which are very expensive to acquire. In this paper, we make the first attempt to train an image captioning model in an unsupervised manner. Instead of relying on manually labeled image-sentence pairs, our proposed model merely requires an image set, a sentence corpus, and an existing visual concept detector. The sentence corpus is used to teach the captioning model how to generate plausible sentences. Meanwhile, the knowledge in the visual concept detector is distilled into the captioning model to guide the model to recognize the visual concepts in an image. In order to further encourage the generated captions to be semantically consistent with the image, the image and caption are projected into a common latent space so that they can reconstruct each other. Given that the existing sentence corpora are mainly designed for linguistic research and are thus with little reference to image contents, we crawl a large-scale image description corpus of two million natural sentences to facilitate the unsupervised image captioning scenario. Experimental results show that our proposed model is able to produce quite promising results without any caption annotations.
\end{abstract}

\section{Introduction}
\begin{figure}
  \centering
  \includegraphics[width=\columnwidth]{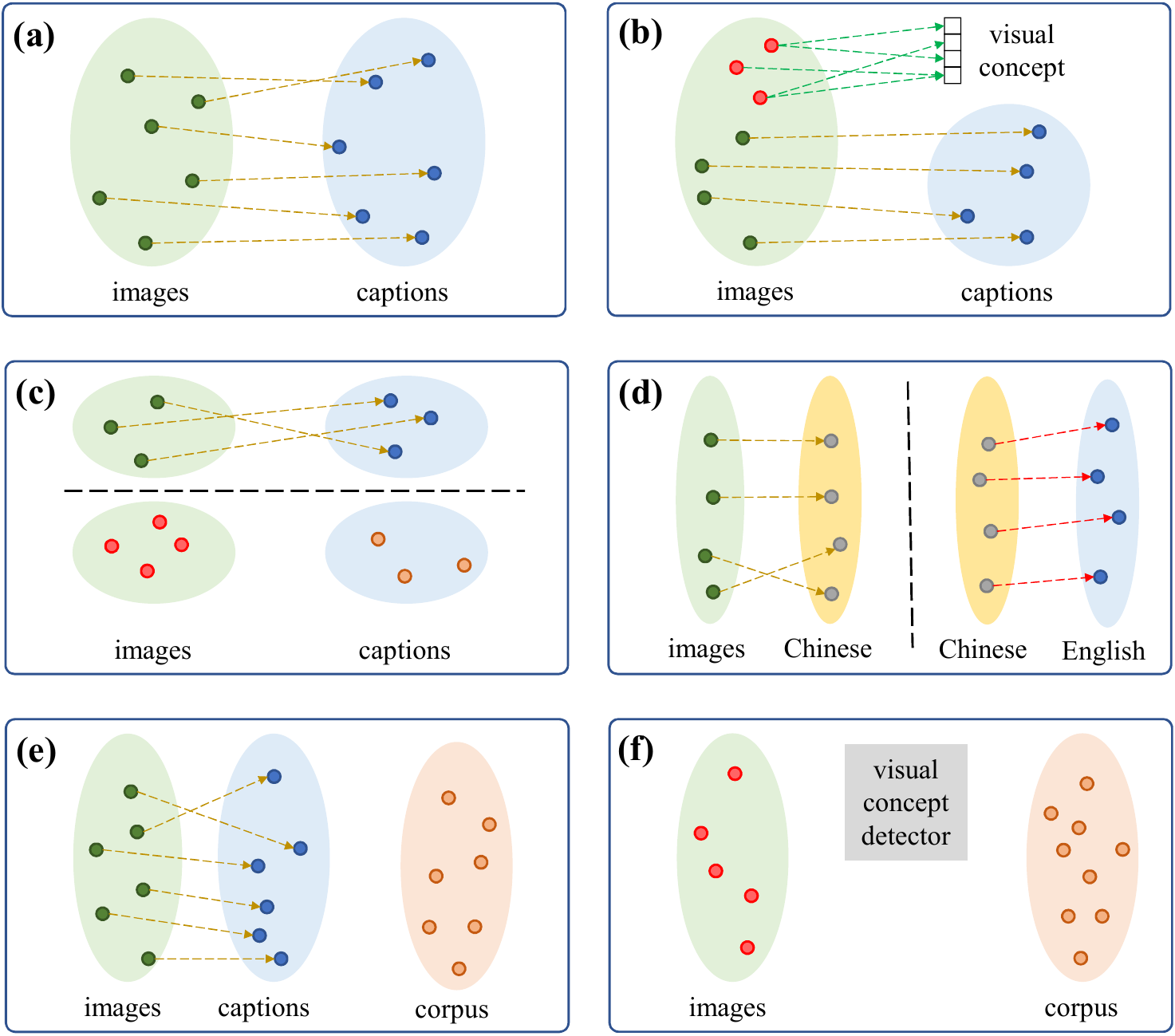}
  \caption{Conceptual differences between the existing captioning methods: (a) supervised captioning~\cite{vinyals2015show}, (b) novel object captioning~\cite{anderson2018partially,anne2016deep}, (c) cross-domain captioning~\cite{chen2017show,zhao2017dual}, (d) pivot captioning~\cite{gu2018unpaired}, (e) semi-supervised captioning~\cite{chen2016semi}, and (f) our proposed unsupervised captioning.}
  \label{fig:1}
\vspace{-0.4cm}
\end{figure}

The research on image captioning has made an impressive progress in the past few years. Most of the proposed methods learn a deep neural network model to generate captions conditioned on an input image~\cite{chen2017sca,chen2018regularizing,gan2017semantic,jiang2018learning,jiang2018recurrent,vinyals2015show,yao2017boosting,you2016image}. These models are trained in a supervised learning manner based on manually labeled image-sentence pairs, as illustrated in Figure \ref{fig:1}~(a). However, the acquisition of these paired image-sentence data is a labor intensive process. The scales of existing image captioning datasets, such as Microsoft COCO~\cite{lin2014microsoft}, are relatively small compared with image recognition datasets, such as ImageNet~\cite{ILSVRC15} and OpenImages~\cite{openimages}. The image and sentence varieties within these image captioning datasets are limited to be under 100 object categories. As a result, it is difficult for the captioning models trained on such paired image-sentence data to generalize to images in the wild~\cite{tran2016rich}. Therefore, how to relieve the dependency on the paired captioning datasets and make use of other available data annotations to well generalize image captioning models is becoming increasingly important, and thus warrants deep investigations.

Recently, there have been several attempts at relaxing the reliance on paired image-sentence data for image captioning training. As shown in Figure~\ref{fig:1} (b), Hendricks \textit{et al.}~\cite{anne2016deep} proposed to  generate captions for novel objects, which are not present in the paired image-caption training data but exist in image recognition datasets, \textit{e.g.}, ImageNet. As such, novel object information can be introduced into the generated captioning sentence without additional paired image-sentence data. A thread of work \cite{chen2017show,zhao2017dual} proposed to transfer and generalize the knowledge learned in existing paired image-sentence datasets to a new domain, where only unpaired data is available, as shown in Figure~\ref{fig:1} (c).  In this way, no paired image-sentence data is needed for training a new image captioning model in the target domain. Recently, as shown in Figure~\ref{fig:1} (d), Gu \textit{et al.}~\cite{gu2018unpaired} proposed to generate captions in a pivot language (Chinese) and then translate the pivot language captions to the target language (English), which requires no more paired data of images and target language captions. Chen \textit{et al.}~\cite{chen2016semi} proposed a semi-supervised framework for image captioning, which uses an external text corpus, shown in  Figure~\ref{fig:1} (d), to pre-train their image captioning model.  Although these methods have achieved improved results, a certain amount of paired image-sentence data is indispensable for training the image captioning models.

To the best of our knowledge, no work has explored unsupervised image captioning, \textit{i.e.}, training an image captioning model without using any labeled image-sentence pairs. 
Figure \ref{fig:1}~(f) shows this new scenario, where only one image set and one external sentence corpus are used in an unsupervised training setting, which, if successful, can dramatically reduce the labeling work required to create a paired image-sentence dataset. However, it is very challenging to figure out how we can leverage the independent image set and sentence corpus to train a reliable image captioning model.

Recently, several models, relying on only monolingual corpora, have been proposed for unsupervised neural machine translation~\cite{artetxe2017unsupervised,lample2017unsupervised}. The key idea of these methods is to map the source and target languages into a common space by a shared encoder with cross-lingual embeddings.
Compared with unsupervised machine translation, unsupervised image captioning is even more challenging. The images and sentences reside in two modalities with significantly different characteristics. Convolutional neural network (CNN)~\cite{lecun1998gradient}  usually acts as an image encoder, while recurrent neural network (RNN)~\cite{hochreiter1997long} is naturally suitable for encoding sentences. Due to their different structures and characteristics, the encoders of image and sentence cannot be shared, as in  unsupervised machine translation. 

In this paper, we make the first attempt to train image captioning models without any labeled image-sentence pairs. Specifically, three key objectives are proposed. First, we train a language model on the sentence corpus using the adversarial text generation method~\cite{fedus2018maskgan}, which generates a sentence conditioned on a given image feature. As illustrated in Figure~\ref{fig:1}~(f), we do not have the ground-truth caption of a training image in the unsupervised setting. Therefore, we employ adversarial training~\cite{goodfellow2014generative} to generate sentences such that they are indistinguishable from the sentences within the corpus.
Second, in order to ensure that the generated captions contain the visual concepts in the image, we distill~\cite{hinton2015distilling} the knowledge provided by a visual concept detector into the image captioning model. Specifically, a reward will be given when a word, which corresponds to the detected visual concepts in the image, appears in the generated sentence. 
Third, to encourage the generated captions to be semantically consistent with the image, the image and sentence are projected into a common latent space. Given a projected image feature, we can decode a caption, which can further be used to reconstruct the image feature. Similarly, we can encode a sentence from the corpus to the latent space feature and thereafter reconstruct the sentence. 
By performing {\it bi-directional reconstructions}, the generated sentence is forced to closely represent the semantic meaning of the image, in turn improving the image captioning model. 

Moreover, we develop an image captioning model initialization pipeline to overcome the difficulties in training from scratch. We first take the concept words in a sentence as input and train a concept-to-sentence model using the sentence corpus only. Next, we use the visual concept detector to recognize the visual concepts present in an image. Integrating these two components together, we are able to generate a pseudo caption for each training image. The {\it pseudo image-sentence pairs} are used to train a caption generation model in the standard supervised manner, which then serves as an initialization for our image captioning model.

In summary, our contributions are four-fold:
\setlist{nolistsep}
\begin{itemize}[noitemsep]
	\item We make the first attempt to conduct unsupervised image captioning without relying on any labeled image-sentence pairs.
    \item We propose three objectives to train the image captioning model. 
    \item We propose a novel model initialization pipeline exploiting unlabeled data. By leveraging the visual concept detector, we generate a pseudo caption for each image and initialize the image captioning model using the pseudo image-sentence pairs.
    \item We crawl a large-scale image description corpus consisting of over two million sentences from the Web for the unsupervised image captioning task. Our experimental results demonstrate the effectiveness of our proposed model in producing quite promising image captions. 
\end{itemize}

\begin{figure*}
  \centering
  \includegraphics[width=\textwidth]{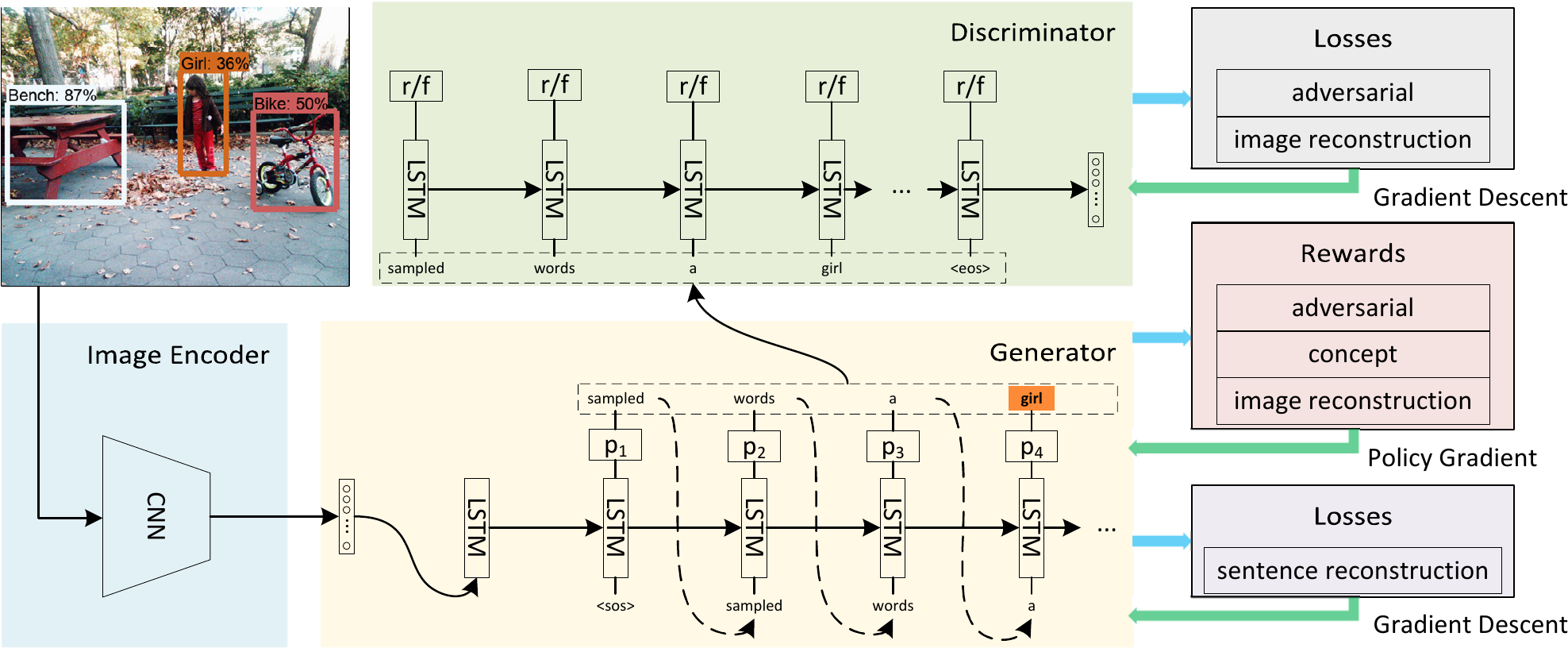}
  \caption{The architecture of our unsupervised image captioning model, consisting of an image encoder, a sentence generator, and a discriminator. A CNN encodes a given image into a feature representation, based on which the generator outputs a sentence to describe the image. The discriminator is used to distinguish whether a caption is generated by the model or from the sentence corpus. Moreover, the generator and discriminator are coupled in a different order to perform image and sentence reconstructions. The adversarial reward, concept reward
, and image reconstruction reward are jointly introduced to train the generator via policy gradient. Meanwhile, the generator is also updated by gradient descent to minimize the sentence reconstruction loss. For the discriminator, its parameters are updated by the adversarial loss and image reconstruction loss via gradient descent.}
  \label{framework}
\vspace{-0.4cm}
\end{figure*}

\section{Related Work}
\subsection{Image Captioning}
Supervised image captioning has been extensively studied in the past few years. Most of the proposed models use one CNN to encode an image and one RNN to generate a sentence describing the image~\cite{vinyals2015show}, respectively. These models are trained to maximize the probability of generating the ground-truth caption conditioned on the input image. As paired image-sentence data is expensive to collect, some researchers tried to leverage other data available to improve the performances of image captioning models.
Anderson \textit{et al.}~\cite{anderson2018partially} trained an image caption model with partial supervision. Incomplete training sequences are represented by finite state automaton, which can be used to sample complete sentences for training.
Chen \textit{et al.}~\cite{chen2017show} developed an adversarial training procedure to leverage unpaired data in the target domain. Although improved results have been obtained, the novel object captioning or domain adaptation methods still need paired image-sentence data for training. Gu \textit{et al.} \cite{gu2018unpaired} proposed to first generate captions in a pivot language and then translate the pivot language caption to the target language. Although no image and target language caption pairs are used, their method depends on image-pivot pairs and a pivot-target parallel translation corpus. In contrast to the methods aforementioned, our proposed method does not need any paired image-sentence data.

\subsection{Unsupervised Machine Translation}
Unsupervised image captioning is similar in spirit to unsupervised machine translation, if we regard the image as the source language. In the unsupervised machine translation methods \cite{artetxe2017unsupervised,lample2017unsupervised,lample2018phrase}, the source language and target language are mapped into a common latent space so that the sentences of the same semantic meanings in different languages can be well aligned and the following translation can thus be performed.
However, the unsupervised image captioning task is more challenging because images and sentences reside in two modalities with significantly different characteristics.

\section{Unsupervised Image Captioning}
\label{method}

Unsupervised image captioning relies on a set of images $\mathcal{I}=\{\bm{I}_1,\ldots,\bm{I}_{N_i}\}$, a set of sentences $\mathcal{\hat{S}}=\{\hat{\bm{S}}_1,\ldots,\hat{\bm{S}}_{N_s}\}$, and an existing visual concept detector, where $N_i$ and $N_s$ are the total numbers of images and sentences, respectively. Please note that the sentences are obtained from an external corpus, which is not related to the images. For simplicity, we will omit the subscripts and use $\bm{I}$ and $\hat{\bm{S}}$ to represent an image and a sentence, respectively. In the following, we first describe the architecture of our image captioning model. Afterwards, we will introduce how to perform the training based on the given data.

\subsection{The Model}
\label{sec:model}
As shown in Figure~\ref{framework}, our proposed image captioning model consists of an image encoder, a sentence generator, and a sentence discriminator.

\textbf{Encoder.} One image CNN encodes the input image into one feature representation $\bm{f}_{im}$:
\begin{equation}
\label{eq:imageCNN}
\bm{f}_{im}=\text{CNN}(\bm{I}).
\end{equation}
Common image encoders, such as Inception-ResNet-v2~\cite{szegedy2017inception} and ResNet-50~\cite{he2016deep}, can be used here. In this paper, we simply choose Inception-V4~\cite{szegedy2017inception} as the encoder.

\textbf{Generator.} Long short-term memory (LSTM), acting as the generator, decodes the obtained image representation into a natural sentence to describe the image content. At each time-step, the LSTM outputs a probability distribution over all the words in the vocabulary conditioned on the image feature and previously generated words. The generated word is sampled from the vocabulary according to the obtained probability distribution:
\begin{equation}
\label{eq:gen}
\begin{aligned}
\bm{x}_{-1}&=\text{FC}(\bm{f}_{im}),\\
\bm{x}_t &= \bm{W}_e\bm{s}_t,\ t\in\{0\ldots n-1\},\\
[\bm{p}_{t+1}, \bm{h}^g_{t+1}] &= \text{LSTM}^g(\bm{x}_t, \bm{h}^g_t),\ t\in\{-1\ldots n-1\},\\
\bm{s}_{t} &\sim \bm{p}_{t},\ t\in\{1\ldots n\},
\end{aligned}
\end{equation}
where $\text{FC}$ and $\sim$ denote the fully-connected layer and sampling operation, respectively. $n$ is the length of the generated sentence with $\bm{W}_e$ denoting the word embedding matrix. $\bm{x}_t$, $\bm{s}_t$, $\bm{h}^g_t$, and $\bm{p}_t$ are the LSTM input, a one-hot vector representation of the generated word, the LSTM hidden state, and the probability over the dictionary at the $t$-th time step, respectively. $\bm{s}_0$ and $\bm{s}_n$ denote the start-of-sentence (SOS) and end-of-sentence (EOS) tokens, respectively. $\bm{h}_{-1}^g$ is initialized with zero. For unsupervised image captioning, the image is not accompanied by sentences describing its content. Therefore, one key difference between our generator and the sentence generator in~\cite{vinyals2015show} is that $\bm{s}_t$ is sampled from the probability distribution $\bm{p}_t$, while the LSTM input word is from the ground-truth caption during training in~\cite{vinyals2015show}.

\textbf{Discriminator.} The discriminator is also implemented as an LSTM, which tries to distinguish whether a partial sentence is a real sentence from the corpus or is generated by the model:
\begin{equation}
\label{eq:dis}
\begin{aligned}
\left[q_t, \bm{h}^d_t\right] &= \text{LSTM}^d(\bm{x}_t, \bm{h}^d_{t-1}),\ t\in\{1\ldots n\},\\
\end{aligned}
\end{equation}
where $\bm{h}^d_t$ is the hidden state of the LSTM. $q_t$ indicates the probability that the generated partial sentence $\bm{S}_t=[\bm{s}_1\ldots\bm{s}_t]$ is regarded as real by the discriminator. Similarly, given a real sentence $\hat{\bm{S}}$ from the corpus, the discriminator outputs $\hat{q}_t, t\in\{1,\ldots,l\}$, where $l$ is the length of $\hat{\bm{S}}$. $\hat{q}_t$ is the probability that the partial sentence with the first $t$ words in $\hat{\bm{S}}$ is deemed as real by the discriminator.

\subsection{Training}
As we do not have any paired image-sentence data available, we cannot train our model in the supervised learning manner. In this paper, we define three novel objectives to make unsupervised image captioning possible.

\begin{figure*}
  \centering
  \includegraphics[width=0.9\textwidth]{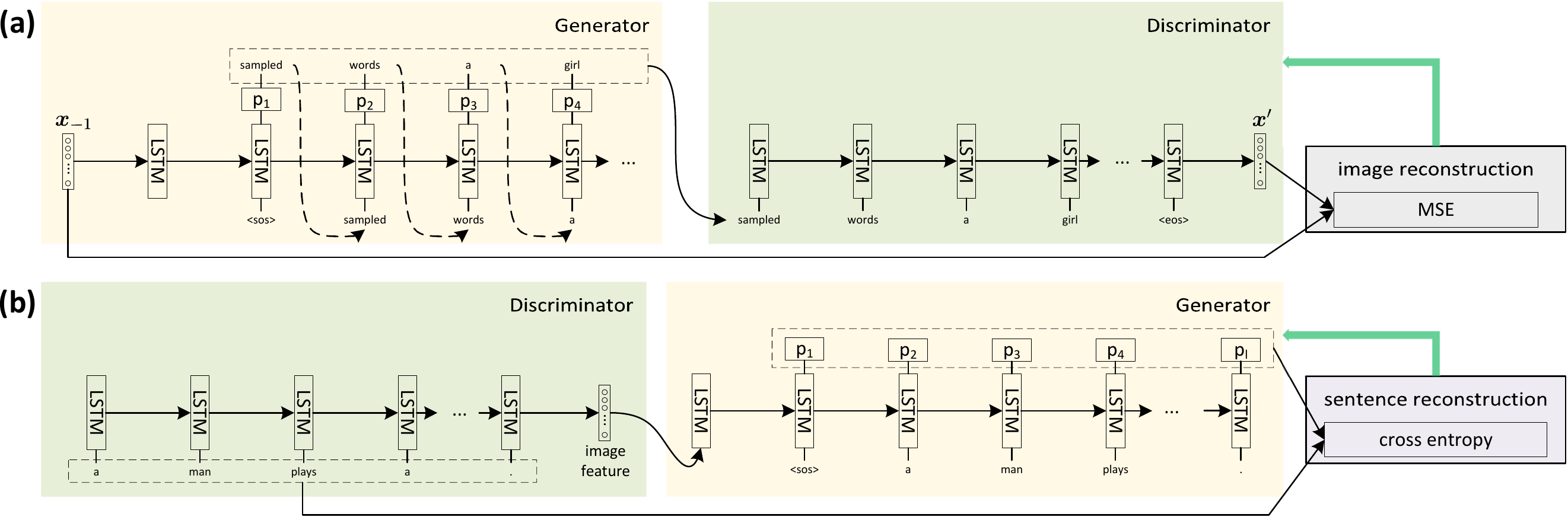}
  \caption{The architectures for image reconstruction (a) and sentence reconstruction (b), respectively, with the generator and discriminator coupled in a different order.}
  \label{fig:dae}
\vspace{-0.5cm}
\end{figure*}

\subsubsection{Adversarial Caption Generation}
\label{sec:obj2}
The sentences generated by the image captioning model need to be plausible to human readers. Such a goal is usually ensured by training a language model on a sentence corpus. However, as discussed before, the supervised learning approaches cannot be used to train the language model in our setting. Inspired by the recent success of the adversarial text generation method \cite{fedus2018maskgan}, we employ the adversarial training~\cite{goodfellow2014generative} to ensure the plausible sentence generation. The generator takes an image feature as input and generates one sentence conditioned on the image feature. The discriminator distinguishes whether a sentence is generated by the model or is a real sentence from the  corpus. The generator tries to fool the discriminator by generating sentences as real as possible. To achieve this goal, we give the generator a reward at each time-step and name this reward as \emph{adversarial reward}. The reward value for the $t$-th generated word is the logarithm of the probability estimated by the discriminator:
\begin{equation}
r_t^{adv} = \log (q_t).
\end{equation}
By maximizing the adversarial reward, the generator gradually learns to generate plausible sentences. For the discriminator, the corresponding adversarial loss is defined as:
\begin{equation}
\mathcal{L}_{adv} =-\left[ \frac{1}{l}\sum_{t=1}^l\log(\hat{q}_t)+\frac{1}{n}\sum_{t=1}^n\log(1-q_t)\right].
\end{equation}

\subsubsection{Visual Concept Distillation}
\label{sec:obj1}
The adversarial reward only encourages the model to generate plausible sentences following grammar rules, which may be irrelevant to the input image. In order to generate relevant image captions, 
the captioning model must learn to recognize the visual concepts in the image and incorporate such concepts in the generated sentence.
Therefore, we propose to distill the knowledge from an existing visual concept detector into the image captioning model. Specifically, when the image captioning model generates a word whose corresponding visual concept is detected in the input image, we give a reward to the generated word. Such a reward is called a  \emph{concept reward}, with the reward value indicated by the confidence score of that visual concept. For an image $\bm{I}$, the visual concept detector outputs a set of concepts and corresponding confidence scores: $\mathcal{C}=\{(c_1,v_1),\ldots,(c_i,v_i),\ldots,(c_{N_c},v_{N_c})\}$, where $c_i$ is the $i$-th detected visual concept, $v_i$ is the corresponding confidence score, and $N_c$ is the total number of detected visual concepts. The \emph{concept reward} assigned to the $t$-th generated word $s_t$ is given by:
\begin{equation}
\begin{aligned}
r^c_t &= \sum_{i=1}^{N_c} \text{I}(s_t=c_i) * v_i, 
\end{aligned}
\end{equation}
where $\text{I}(\cdot)$ is the indicator function. 


\subsubsection{Bi-directional Image-Sentence Reconstruction}
\label{sec:obj31}
With the adversarial training and concept reward, the captioning quality would be largely determined by the visual concept detector because it is the only bridge between images and sentences. However, the existing visual concept detectors can only reliably detect a limited number of object concepts. 
The image captioning model should understand more semantic concepts of the image for a better generalization ability. To achieve this goal, we propose to project the images and sentences into a common latent space such that they can be used to reconstruct each other.
Consequently, the generated caption would be semantically consistent with the image.



\textbf{Image Reconstruction.} The generator produces a sentence conditioned on an image feature, as shown in Figure~\ref{fig:dae}~(a). The sentence caption should contain the gist of the image. Therefore, we can reconstruct the image from the generated sentence, which can encourage the generated captions to be semantically consistent with the image. However, one hurdle for doing so lies in that it is very difficult to generate images containing complex objects, \textit{e.g.}, people, of high-resolution using current techniques \cite{brock2018large,karras2017progressive}. Therefore, in this paper, we turn to {\it reconstruct the image features} instead of the full image. 
As shown in Figure~\ref{fig:dae}~(a), the discriminator can be viewed as a sentence encoder.
A fully-connected layer is stacked on the discriminator to project the last hidden state $\bm{h}_n^d$ to the common latent space for images and sentences:
\begin{equation}
\label{eq:fc}
\bm{x}'=\text{FC}(\bm{h}_n^d),
\end{equation}
where $\bm{x}'$ can be further viewed as the reconstructed image feature from the generated sentence. Therefore, we define an additional image reconstruction loss for training the discriminator:
\begin{equation}
\mathcal{L}_{im} = \|\bm{x}_{-1} - \bm{x}'\|_2^2.
\end{equation}
It can also be observed that the generator together with the discriminator constitutes the image reconstruction process. Therefore, an \textit{image reconstruction reward} for the generator, which is proportional to the negative reconstruction error, can be defined as:
\begin{equation}
r_t^{im} = -\mathcal{L}_{im}.
\end{equation}

\textbf{Sentence Reconstruction.} Similarly, as shown in Figure~\ref{fig:dae}~(b), the discriminator can encode one sentence and project it into the common latent space, which can be viewed as one image representation related to the given sentence. The generator can reconstruct the sentence based on the obtained representation. Such a sentence reconstruction process could also be viewed as a sentence denoising auto-encoder~\cite{vincent2008extracting}. Besides aligning the images and sentences in the latent space, it also learns how to decode a sentence from an image representation in the common space.
In order to make a reliable and robust sentence reconstruction, we add noises to the input sentences by  following~\cite{lample2017unsupervised}.
The objective of the sentence reconstruction is defined as the cross-entropy loss:
\begin{equation}
\mathcal{L}_{sen} = -\sum_{t=1}^l \log\big(p(s_t=\hat{s}_t|\hat{s}_1,\ldots,\hat{s}_{t-1})\big),
\end{equation}
where $\hat{s}_t$ is the $t$-th word in sentence $\hat{\bm{S}}$.

\subsubsection{Integration}
The three objectives are jointly considered to train our image captioning model. For the generator, as the word sampling operation is not differentiable, we train the generator using policy gradient~\cite{sutton2000policy}, which estimates the gradients with respect to trainable parameters given the joint reward. More specifically, the joint reward consists of \textit{adversarial reward}, \textit{concept reward}, and \textit{image reconstruction reward}. Besides the gradients estimated by policy gradient, the sentence reconstruction loss also provides gradients for the generator via back-propagation. These two types of gradients are both employed to update the generator. Let $\theta$ denote the trainable parameters in the generator. The gradient with respect to $\theta$ is given by:
\begin{equation}
\begin{aligned}
&\nabla_\theta \mathcal{L}(\theta) =-\mathbb{E}\left[\sum_{t=1}^n\left(\sum_{s=t}^n\gamma^s(
\underbrace{r_s^{adv}}_{\text{adversarial}}+
\underbrace{\lambda_cr_s^c}_{\text{concept}}) \right.\right.\\ & \left.\left.  +
\underbrace{\lambda_{im}r_s^{im}}_{\text{image reconstruction}}-b_t\right)\nabla_\theta \log(\bm{s}_t^\top\bm{p}_t)\right]+
\underbrace{\lambda_{sen}\nabla_\theta\mathcal{L}_{sen}(\theta)}_{\substack{\text{sentence} \\ \text{reconstruction}}},
\end{aligned}
\end{equation}
where $\gamma$ is a decay factor, and $b_t$ is the baseline reward estimated using self-critic \cite{rennie2017self}. $\lambda_c$, $\lambda_{im}$ and $\lambda_{sen}$ are the hyper-parameters controlling the weights of different terms.

For the discriminator, the adversarial and image reconstruction losses are combined to update the parameters via gradient descent:
\begin{equation}
\mathcal{L}_D=\mathcal{L}_{adv}+\lambda_{im}\mathcal{L}_{im}.
\end{equation}
During our training process, the generator and discriminator are updated alternatively. 

\subsection{Initialization}
\label{sec:init}
It is challenging to adequately train our image captioning model from scratch with the given unlabeled data, even with the proposed three objectives. Therefore, we propose an initialization pipeline to pre-train the generator and discriminator.


Regarding the generator, we would like to generate a pseudo caption for each training image, and then use the pseudo image-caption pairs to initialize an image captioning model. Specifically, we first build a concept dictionary consisting of the object classes in the OpenImages dataset~\cite{openimages}. Second, we train a concept-to-sentence (con2sen) model using the sentence corpus only. Given a sentence, we use a one-layer LSTM to encode the concept words within the sentence into a feature representation, and use another one-layer LSTM to decode the representation into the whole sentence.
Third, we detect the concepts of each image by the existing visual concept detector. With the detected concepts and the concept-to-sentence model, we are able to generate a pseudo caption for each image.
Fourth, we train the generator with the pseudo image-caption pairs using the standard supervised learning method as in~\cite{vinyals2015show}. Such an image captioner is named as {feature-to-sentence} (feat2sen) and used to initialize the generator.

Regarding the discriminator, parameters are initialized by training an adversarial sentence generation model on the sentence corpus.

\section{Experiments}
In this section, we evaluate the effectiveness of our proposed method. To quantitatively evaluate our unsupervised captioning method, we use the images in the MSCOCO dataset~\cite{lin2014microsoft} as the image set (excluding the captions). The sentence corpus is collected by crawling the image descriptions from Shutterstock\footnote{\url{https://www.shutterstock.com}}. The object detection model~\cite{huang2017speed} trained on OpenImages~\cite{openimages} is used as the visual concept detector. We first introduce sentence corpus crawling and experimental settings. Next, we present the performance comparisons as well as the ablation studies.

\begin{figure}
\centering
\includegraphics[width=0.8\columnwidth]{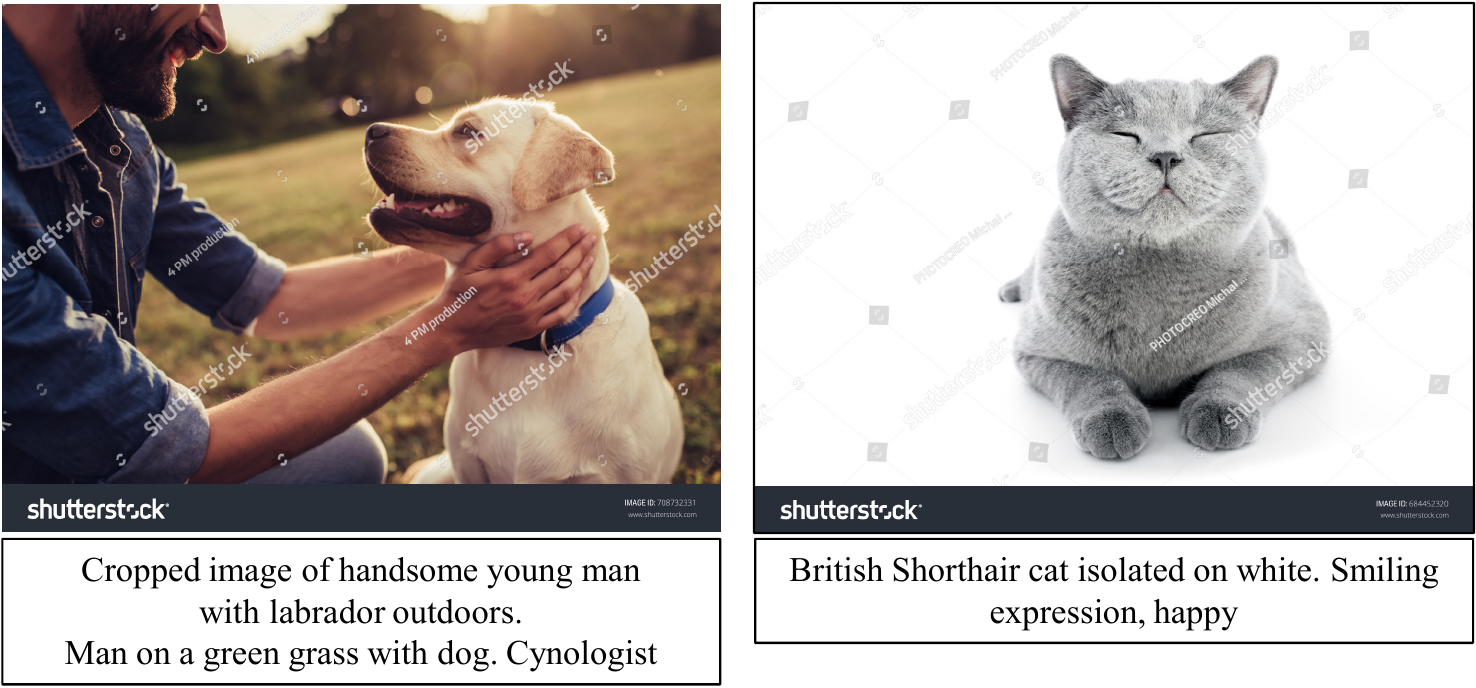}
\caption{\label{fig:shutter} Two images and their accompanying descriptions from Shutterstock.}
\vspace{-0.4cm}
\end{figure}

\subsection{Shutterstock Image Description Corpus}
We collect a sentence corpus by crawling the image descriptions from Shutterstock for the unsupervised image captioning research. Shutterstock is an online stock photography website, which provides hundreds of millions of royalty-free stock images. Each image is uploaded with a description written by the image composer. Some images and description samples are shown in Figure~\ref{fig:shutter}. 
We hope the crawled image descriptions to be somewhat related to the training images. Therefore, we directly use the name of the eighty object categories in the MSCOCO dataset as the search keywords. For each keyword, we download the search results of the top one thousand pages. If the number of pages available is less than one thousand, we will download all the results. There are roughly a hundred images in one page, resulting in $100,000$ descriptions for each object category. After removing the sentences with less than eight words, we collect $2,322,628$ distinct image descriptions in total.

\begin{figure*}
  \centering
  \includegraphics[width=\textwidth]{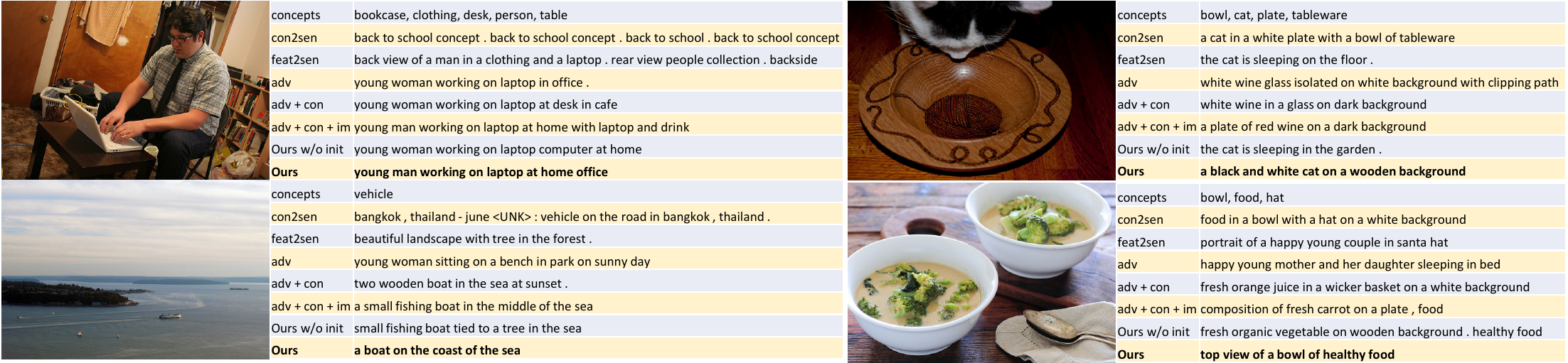}
  \caption{The qualitative results by the unsupervised captioning methods trained with different objectives. 
  Best viewed by zooming in.}
  \label{qualitative}
\vspace{-0.4cm}
\end{figure*}

\subsection{Experimental Settings}
Following~\cite{karpathy2015deep}, we split the MSCOCO dataset, with 113,287 images for training, 5,000 images for validation, and the remaining 5,000 images for testing. Please note that the training images are used to build the image set, {\it with the corresponding captions left unused for any training}. All the descriptions in the Shutterstock image description corpus are tokenized by the NLTK toolbox~\cite{bird2009natural}. We build a vocabulary by counting all the tokenized words and removing the words with frequency lower than 40. 
The object category names of the used object detection model are then merged into the vocabulary. Finally, there are $18,667$ words in our vocabulary, including special SOS, EOS, and an Unkown token. 
We perform a further filtering process by removing the sentences containing more than 15\% Unknown token. After filtering, we retain $2,282,444$ sentences.

The LSTM hidden dimension and the shared latent space dimension are fixed to 512. The weighting hyper-parameters are chosen to make different rewards roughly at the same scale. Specifically, $\lambda_c$, $\lambda_{im}$, $\lambda_{sen}$ are set to be 10, $0.2$, and 1, respectively. $\gamma$ is set to be $0.9$. We train our model using the Adam optimizer~\cite{kingma2014adam} with a learning rate of 0.0001. During the initialization process, we minimize the cross-entropy loss using Adam with the learning rate 0.001. When generating the captions in the test phase, we use beam search with a beam size of 3.

We report the BLEU~\cite{papineni2002bleu}, METEOR~\cite{denkowski2014meteor}, ROUGE~\cite{lin2004rouge}, CIDEr~\cite{vedantam2015cider}, and SPICE~\cite{anderson2016spice} scores computed with the coco-caption code \footnote{\url{https://github.com/tylin/coco-caption}}. The ground-truth captions of the images in the test split are used for computing the evaluation metrics.

\begin{table}
  \caption{Performance comparisons of unsupervised captioning methods on the test split~\cite{karpathy2015deep} of the MSCOCO dataset. 
  }
  \label{tab:results}
  \centering
  \begin{adjustbox}{max width=\columnwidth}
  \begin{tabular}{l|cccccccc}
    \toprule
    Method & B1 & B2 & B3 & B4 & M & R & C & S \\
    \midrule
    Ours w/o init & 38.2 & 20.6 & 9.9 & 4.8 & 11.2 & 27.5 & 22.9 & 6.6 \\
    \rowcolor{mygray}
    Ours & 41.0 & 22.5 & 11.2 & 5.6 & 12.4 & 28.7 & 28.6 & 8.1 \\
    \midrule
    con2sen & 37.2 & 20.0 & 9.6 & 4.7 & 12.3 & 27.3 & 22.5 & 8.2 \\
    feat2sen & 38.7 & 21.3 & 10.3 & 5.0 & 12.4 & 28.3 & 23.5 & 8.0 \\
    adv  & 34.0 & 15.6 & 6.5 & 2.9 & 8.7 & 24.2 & 11.8 & 3.8 \\
    adv + con & 37.9 & 19.8 & 9.4 & 4.6 & 11.4 & 26.5 & 24.1 & 7.3 \\
    adv + con + im & 37.8 & 19.9 & 9.5 & 4.6 & 11.9 & 26.8 & 25.5 & 7.5 \\
    \bottomrule
  \end{tabular}
  \end{adjustbox}
\vspace{-0.4cm}
\end{table}

\subsection{Experimental Results and Analysis}
The top region of Table~\ref{tab:results} illustrates the unsupervised image captioning results on the test split of the MSCOCO dataset. The captioning model obtained with the proposed unsupervised training method achieves promising results, with CIDEr as 28.6\%. Moreover, we also report the results of training our model from scratch (``Ours w/o init'') to verify the effect of our proposed initialization pipeline. Without initialization, the CIDEr value drops to 22.9\%, which shows that the initialization pipeline can benefit the model training and thus boost image captioning performances.



\textbf{Ablation Studies.} The results of the ablation studies are illustrated in the bottom region of Table \ref{tab:results}. 
It can be observed that ``con2sen'' and ``feat2sen'' generate reasonable results with CIDEr as 22.5\% and 23.5\%, respectively. As such, ``con2sen'' can be used to generate pseudo image-caption pairs for training ``feat2sen''. And ``feat2sen'' can make a meaningful initialization of the generator of our captioning model.

When only the adversarial objective is introduced to train the captioning model, ``adv'' alone leads to much worse results. 
One cause for this is due to the linguistic characteristics of the crawled image descriptions from Shutterstock, which is significantly different from that of the COCO caption. Another cause is that the adversarial objective only enforces genuine sentence generation but does not ensure its semantic correlation with the image content. Because of the linguistic characteristic difference, most metrics also drop even after introducing the concept objective in ``adv + con'' and further incorporating image reconstruction objective in ``adv + con + im''. Although the generated sentences of these two baselines may look plausible, the evaluation results with respect to the COCO captions are not satisfactory. However, by considering all the objectives together, our proposed method  substantially improves the captioning performances.

\textbf{Qualitative Results.} Figure \ref{qualitative} shows some qualitative results of unsupervised image captioning. 
In the top-left image, the object detector fails to detect the ``\texttt{laptop}''. So ``con2sen'' model says nothing about the laptop. On the contrary, the other models successfully recognize laptops and incorporate such concepts into the generated caption. In the top-right image, only a small region of the cat is visible. With such a small region, our full captioning model recognizes that it is ``\texttt{a black and white cat}''. The object detector cannot provide any information about color attribute. We are pleased  to see that the bi-directional reconstruction objective is able to guide the captioning model to recognize and express such visual attributes in the generated description sentence.
In the bottom two images, ``\texttt{vehicle}'' and ``\texttt{hat}'' are detected by error, which severely  affects the results of ``con2sen''. On the contrary, after training the captioning model with the proposed objectives, the captioning model is able to correct such errors and generate plausible captions.

\begin{figure}
\vspace{-0.6cm}
\centering
\includegraphics[width=0.8\columnwidth]{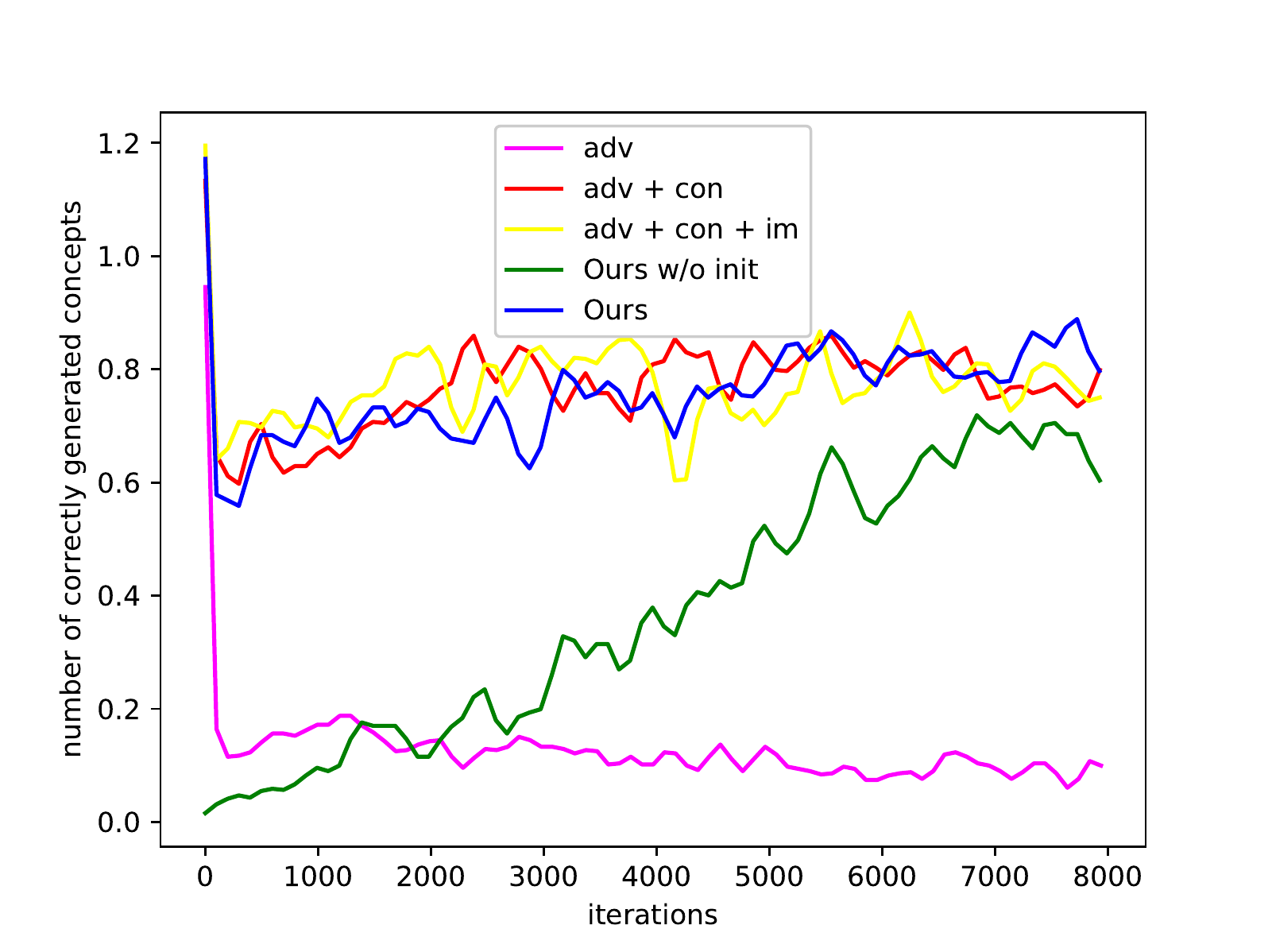}
\caption{\label{fig:obj}The averaged number of correct concept words in each sentence generated during the training process.}
\vspace{-0.5cm}
\end{figure}

\textbf{Effect of Concept Reward.} Figure \ref{fig:obj} shows the averaged number of correct concept words in each sentence generated during the training process. It can be observed that the number of ``adv'' drops quickly in the beginning. The reason is that the adversarial objective is not related to the visual concepts in the image. ``Ours w/o init'' continuously increases from zero to about 0.6. The concept reward consistently improves the ability of the captioning model to recognize visual concepts. For ``adv + con'', ``adv + con + im'', and ``Ours'', the number is about 0.8. One reason is that the initialization pipeline gives a good starting point. Another possible reason is that the concept reward prevents the captioning model from drifting towards degradation.

\subsection{Performance Comparisons under the Unpaired Captioning Setting}
The performance of unsupervised captioning model may seem unsatisfactory in terms of the evaluation metrics on the COCO test split. This is mainly due to the different linguistic characteristics between COCO captions and crawled image descriptions. To further demonstrate the effectiveness of the proposed three objectives, we compare with~\cite{gu2018unpaired} {\it under the same unpaired captioning setting}, where  the COCO captions of the training images are used but in an unpaired manner. Specifically, we replace the crawled sentence corpus with the COCO captions of the training images. All the other settings are kept the same as the unsupervised captioning settings. A new vocabulary with $11,311$ words is created by counting all the words in the training captions and removing the words with frequency less than 4.

The results of unpaired image captioning are shown in Table~\ref{tab:unpaired}. It can be observed that the captioning model can be consistently improved based on the unpaired data, by including the three proposed objectives step by step. Due to exposure bias~\cite{ranzato2015sequence}, some of the captions generated by ``feat2sen'' are poor sentences. The adversarial objective encourages these generated sentences to appear genuine, resulting in improved performances. 
With only adversarial training, the model tends to generate sentences unrelated to the image. This problem is mitigated by the concept reward and thus ``adv + con'' leads to an even better performance. By only including the image reconstruction objective, ``adv + con + im'' provides a minor improvement. However, if we include the sentence reconstruction objective, our full captioning model achieves another significant improvement, with CIDEr value increasing from 49\% to 54.9\%. The reason is that the bi-directional image and sentence reconstruction can further leverage the unpaired data to encourage the generated caption to be semantically consistent with the image. The proposed method obtains significantly better results than~\cite{gu2018unpaired}, which may be because that the information in the COCO captions is more adequately exploited in our proposed method. 

\begin{table}
  \caption{Performance comparisons  on the test split~\cite{karpathy2015deep} of the MSCOCO dataset {\it under the unpaired setting}.}
  \label{tab:unpaired}
  \centering
  \begin{adjustbox}{max width=\columnwidth}
  \begin{tabular}{l|cccccccc}
    \toprule
    Method & B1 & B2 & B3 & B4 & M & R & C & S \\
    \midrule
    Pivoting \cite{gu2018unpaired} & 46.2 & 24.0 & 11.2 & 5.4 & 13.2 & - & 17.7 & - \\
    Ours w/o init & 53.8 & 35.5 & 23.1 & 15.6 & 16.6 & 39.9 & 46.7 & 9.6 \\
    \rowcolor{mygray}
    Ours & 58.9 & 40.3 & 27.0 & 18.6 & 17.9 & 43.1 & 54.9 & 11.1 \\
    \midrule
    con2sen & 50.6 & 30.8 & 18.2 & 11.3 & 15.7 & 37.9 & 33.9 & 9.1 \\
    feat2sen & 51.3 & 31.3 & 18.7 & 11.8 & 15.3 & 38.1 & 35.4 & 8.8 \\
    adv & 55.6 & 35.5 & 23.1 & 15.7 & 17.0 & 40.8 & 45.8 & 10.1 \\
    adv + con & 56.2 & 37.2 & 24.2 & 16.2 & 17.3 & 41.5 &  48.8 & 10.5 \\
    adv + con + im & 56.4 & 37.5 & 24.5 & 16.5 & 17.4 & 41.6 & 49.0 & 10.5 \\
    \bottomrule
  \end{tabular}
  \end{adjustbox}
\vspace{-0.4cm}
\end{table}

\section{Conclusions}
In this paper, we proposed a novel method to train an  image captioning model in an unsupervised manner without using any paired image-sentence data. As far as we know, this is the first attempt to investigate  this problem. To achieve this goal, we presented three training objectives, which encourage that 1) the generated captions are indistinguishable from sentences in the corpus, 2) the image captioning model conveys the object information in the image, and 3) the image and sentence features are aligned in the common latent space and perform bi-directional reconstructions from each other. A large-scale image description corpus consisting of over two million sentences was further collected from Shutterstock to facilitate the unsupervised image captioning method. The experimental results demonstrate that the proposed method can produce quite promising results without using any labeled image-sentence pairs. In the future, we will conduct human evaluations for unsupervised image captioning.

\section*{Acknowledgement}
This work is partially supported by NSF awards \#1704309,  \#1722847, and \#1813709.

{\small
\bibliographystyle{ieee_fullname}
\bibliography{egbib}
}

\end{document}